\documentclass{article}
\usepackage{ijcai19}

\usepackage{times}
\pdfoutput=1
\usepackage{soul}
\usepackage{url}
\usepackage[hidelinks]{hyperref}
\usepackage[utf8]{inputenc}
\usepackage[small]{caption}
\usepackage{graphicx}
\usepackage{amsmath}
\usepackage{booktabs}
\usepackage{algorithm}
\usepackage{algorithmic}
\usepackage{natbib}
\usepackage{amsmath,amssymb,amsfonts}

\renewcommand{\cite}[1]{\citep{#1}}
\DeclareMathOperator*{\argmax}{arg\,max}

\usepackage[colorinlistoftodos]{todonotes}
\usepackage{latexsym} 
\usepackage{caption}
\usepackage{subcaption}
\usepackage{balance}

\title{Inferring Personalized Bayesian Embeddings for Learning from Heterogeneous Demonstration}

\author{
    Rohan Paleja and
    Matthew Gombolay
    \affiliations
    Georgia Institute of Technology, Atlanta, GA, USA \emails
    rpaleja3@gatech.edu, Matthew.Gombolay@cc.gatech.edu
    }

\begin{document}

\maketitle

\begin{abstract}
\label{abstract}
For assistive robots and virtual agents to achieve ubiquity, machines will need to anticipate the needs of their human counterparts. The field of Learning from Demonstration (LfD) has sought to enable machines to infer predictive models of human behavior for autonomous robot control. However, humans exhibit heterogeneity in decision-making, which traditional LfD approaches fail to capture. To overcome this challenge, we propose a Bayesian LfD framework to infer an integrated representation of all human task demonstrators by inferring human-specific embeddings, thereby distilling their unique characteristics. We validate our approach is able to outperform state-of-the-art techniques on both synthetic and real-world data sets. 
\end{abstract}

\section{Introduction}

Human teams develop shared mental models over years of experience that enable them to anticipate each others' action without explicit communication. This anticipatory ability is a key enabler of high-performing teams \cite{doi:10.1177/0149206309356804}. By contrast, it is for the same reason that assistive machines (e.g., ``cobots") have not achieved proliferation. Humans currently must explicitly communicate task directives to machines, placing a substantial burden on the human teammates. The field of Learning from Demonstration (LfD) has sought to enable robot's to infer such models of human behavior to either imitate the process (i.e., apprenticeship learning~\cite{Abbeel:2004:ALV:1015330.1015430} or to enable machines to anticipate what actions humans are likely to take so as to take a supportive action~\cite{nikolaidis2012human}. LfD has achieved success in learning to drive ~\cite{Abbeel:2004:ALV:1015330.1015430}, fly~\cite{coates2009apprenticeship}, and dexterous manipulation~\cite{lee2015learning}.

\begin{figure}[btp]
	\begin{center}
		\centerline{\includegraphics[width = 0.5\textwidth]{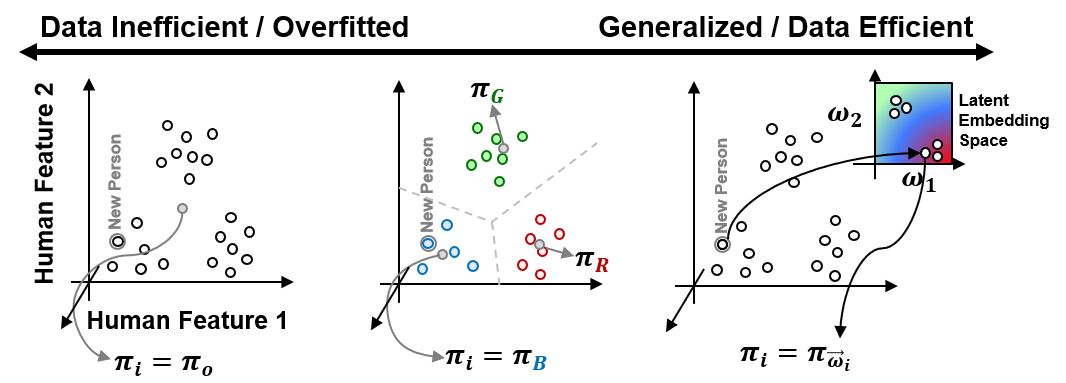}}
		\caption{Depiction of approaches to heterogeneity: (Left) Assume homogeneity~\cite{DBLP:conf/icml/SammutHKM92}, (Center) Partition data to semi-homogeneous clusters~\cite{Nikolaidis:2015:EML:2696454.2696455}, and (Right) Bayesian embeddings (our approach).}
		\label{fig:power_of_BNN}
	\end{center}
\end{figure}

However, prior approaches to LfD struggle to handle heterogeneity in human behavior among the demonstrators. The typical approach to LfD is to assume homogeneity, reasoning about the average human, as shown in the left-most diagram of Figure \ref{fig:power_of_BNN}. A more recent approach by Nikolaidis et al.~performed k-means clustering over the dynamics of the human environment and then learned a separate model of the human for each cluster, as shown in the center diagram of Figure \ref{fig:power_of_BNN}~\cite{Nikolaidis:2015:EML:2696454.2696455}. While a step forward, this approach greatly reduces the total amount of data to train and does not afford leveraging the possible homogeneity that does exist (e.g., adherence to job-specific constraints). 

In this paper, we seek to overcome these key gaps in prior work that either ignore or attempt to eliminate heterogeneity. We propose learning personalized Bayesian embeddings, inferred through Bayesian neural network (BNN) backpropagation. These personalized embeddings enables the model to adapt to a person's unique characteristics while simultaneously leveraging any homogeneity that exists (i.e., uniform adherence to hard constraints). As human demonstration data is often limited in quantity~\cite{amershi2014power}, we formulate the learning problem to adopt counterfactual reasoning via pairwise comparisons between the action a human took relative to all actions she did not take in order to provide additional structure for the learning problem and increase the amount of available data.


We evaluate our approach with a synthetic dataset consisting of mock experts' scheduling heuristics for jobshop scheduling, and a real-world dataset with human gameplay in StarCraft II. We demonstrate the following novel contributions: First, this is the first paper we are aware of to employ BNNs for LfD. Second, we uniquely develop a counterfactual reasoning model with BNNs for imitation learning. Third, we demonstrate new algorithm, which we call the ``Hybrid-Bayes Shift'' algorithm, to allow for robust performance early on while  inferring an accurate Bayesian embedding for long-term, customized behavior prediction. Our results indicate a strong improvement over prior work, e.g., k-means clustering before training.

\section{Related Work}
\label{related-work}
Researchers have made significant progress in the field of Learning from Demonstration \cite{Abbeel:2004:ALV:1015330.1015430, Konidaris11a, 10.1007/978-3-319-19551-3_26, Ziebart:2008:MEI:1620270.1620297, Konidaris12a, chernova2007confidence,Terrell:2012:RAM:2392800.2392849}. While the ability to capture domain-expert knowledge from demonstration has improved, heterogeneity apparent in the data remains a challenge. Seminal work in LfD by researchers, such as \citet{DBLP:conf/icml/SammutHKM92} found that demonstrator heterogeneity represented a significant challenge to learning from various data sources. In their work, Sammut et. al. found that pilots executing the same flight plan created such variance in the data as to make it more practical to learn from a single trajectory and disregard the remaining data. Rather than reduce the utility of our inference model by fitting it to one data point of many, it would be beneficial to reason about the data set and utilize it in its entirety.

To cope with heterogeneity, more recent work by \citet{Nikolaidis:2015:EML:2696454.2696455} performed unsupervised clustering on the data and learned a model for each cluster. The clustering sought to break the dataset into smaller sets that each contain less variance and thus avoided the difficulty faced by \citet{DBLP:conf/icml/SammutHKM92}. There are several caveats of this approach; each model received a much smaller set of data to learn from and thus the model may not have as much representational power as if it were to learn from the entire set. Secondly, given a new data point, this approach required the use of one of the $k$ models (where $k$ is the number of clusters) for prediction and did not fully account for cases where the human demonstrator is a combination of the clusters. Lastly, this approach did not account for variability within a cluster. Personalizing a model to adhere to a specific person rather than utilizing a group model can result in large performance gains. We note that the closest work to ours is by \citet{DBLP:conf/aaai/KillianKD17}. However, these researchers consider only a synthetic dataset and only model transition dynamics rather than decision-making behavior \`{a} la LfD.


\section{Preliminaries}

\newcommand{\tuple}[1]{\ensuremath{\left \langle #1 \right \rangle }}

LfD mechanisms are typically based upon a Markov Decision Process (MDP). Formally, a MDP is a 5-tuple $M = \langle S,A,T,\gamma,R \rangle $ where $S$ is a set of states, $A$ is a set of actions, $T: S \times A \times S \rightarrow [0,1]$ is a transition function, where $T(s,a,s')$ is the probability of being in state $s'$ after executing action $a$ in state $s$, $R$: $S \rightarrow \mathbb{R}$ (or $R:S \times A \rightarrow \mathbb{R}$) is a reward function that takes the form of $R(s)$ or $R(s,a)$ depending upon whether the reward is assessed for being in a state or for taking a particular action within a state, and $\gamma \in [0,1)$ is the discount factor for future rewards. The goal is then to learn a policy, $\pi:S\rightarrow A$ to maximize the policy's future expected reward, $V^{\pi}(s) = \mathbb{E}[\sum_{t=0}^{\infty} \gamma^t r_t | s_o = s$], where the policy begins in state $s_o = s$ and follows policy $\pi$ thereafter, receiving reward, $r_t$, at each time step. 

The problem of LfD is to receive 1) a set of trajectories provided by a human demonstrator, $\{\tuple{s_t,a_t}, t \in \{1,2,\ldots\}\}$ and 2) an MDP sans reward function, $R$, and recover a policy that can predict the correct state-action sequence a human would take in a novel situation which would maximize the human's latent reward function. There are two basic approaches: goal-based reasoning and policy-based reasoning. Goal-based reasoning captures the work of the Inverse Reinforcement Learning Community~\cite{Abbeel:2004:ALV:1015330.1015430}, in which one attempts to infer the latent reward function (i.e., a set of weights over hand-specified feature counts) and then construct a policy to maximize this reward function. On the other hand, policy-based reasoning seeks to directly learn a mapping from states to actions~\cite{chernova2007confidence}. Goal-based reasoning approaches typically require one to assume a reward basis function (e.g., a weighted, linear combination of feature counts)~\cite{Abbeel:2004:ALV:1015330.1015430} or a Bayesian prior over hand-specified goals~\cite{ziebart2008maximum}. In this work, we desire to have a model that does not require one to have access to such information; as such, we develop a policy learning-based approach.

\section{BNNs for Human Control Policy Modeling}
In this section, we present our model for inferring personalized embeddings to capture the the homo- and heterogeneity among human demonstrators in LfD tasks. This formulation is the first we know of for explicitly reasoning about heterogeneity with a single, integrated model in LfD. 

\begin{figure}[ht]
\begin{center}
\centerline{\includegraphics[width =\columnwidth]{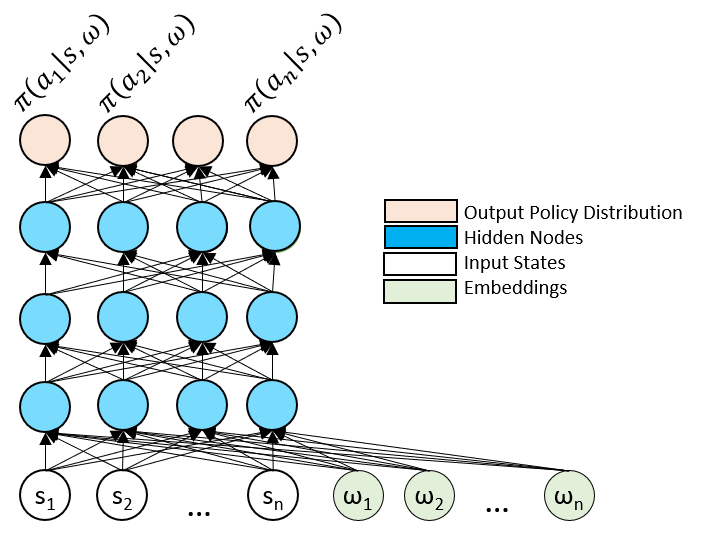}}
\caption{Bayesian Neural Network Diagram.}
\label{BNN}
\end{center}
\end{figure}

Figure \ref{BNN} depicts a BNN, which learns a model, $\pi_\theta : S \times \Omega \to [0,1]^{|A|}$, of the human demonstrator's policy, where $\omega\in\mathbb{R}^d$ is the demonstrator-specific Bayesian embedding of length $d$, which is a tunable hyperparameter. These latent features, $\omega$, provide the ``style" of the current demonstrator, which accounts for a component not represented within the state features and that is needed for accurate prediction. The training procedure of a BNN consists of taking as input an example of a state, $s_t^p$ at time $t$, for person, $p$, as well as the person's embedding, $\omega_p^{(i)}$ at training iteration $i$, and predict the person's action in that state, $y_t^p$. The loss is computed as the R\'{e}nyi divergence~\cite{zyczkowski2003renyi} between the predicted action, $\hat{y}_t^p$, and the true action, $y_t^p$, as shown in Equation \ref{eq:alpha}, where $\vec{e}_{y_{t,j}^p}$ is a one-hot encoding of the demonstrator's action, and subscript $j$ corresponds to the probability for action $j\in{1,2,\ldots,|A|}$. $\alpha \subset (-\infty, +\infty)$ corresponds to a hyperparameter that can be tuned in order to learn a closer approximate distribution to $y_t^p$.
\begin{equation}
D_{\alpha}(\hat{y}_t^p \| \vec{e}_{y_t^p}) = \frac{1}{\alpha-1} \log \left(\sum_{j=1}^{|A|} \frac{\left(\hat{y}_{t,j}^p\right)^\alpha}{\left(\vec{e}_{y_{t,j}^p}\right)^{\alpha-1}}\right)
\label{eq:alpha}
\end{equation}
This loss is then backpropagated through the network to update model parameters $\theta$ and the personalized embedding $\omega$ via stochastic gradient descent as shown in Equations \ref{eq:BP_theta} and \ref{eq:BP_omega}, respectively, at iteration $i$.
\begin{align}
\theta^{(i+1)} &\leftarrow \theta^{(i)} + \alpha_\theta \frac{\partial D_\alpha(\hat{y}_t^p \| \vec{e}_{y_t^p})}{\partial \theta}
\label{eq:BP_theta}\\
\omega^{(i+1)} &\leftarrow \omega^{(i)} + \alpha_\omega \frac{\partial D_\alpha(\hat{y}_t^p \| \vec{e}_{y_t^p})}{\partial \omega} 
\label{eq:BP_omega}
\end{align}
When applying the algorithm during runtime (i.e., testing) for a new human demonstrator, $p'$, one updates the embedding, $\omega_{p'}$ via Equation \ref{eq:BP_omega}; however, the network's parameters, $\theta$, remain static. This hybrid approach enables one to balance the bias-variance tradeoff, grounding the model in parameters common to all demonstrators via $\theta$ while tailoring a subset of the parameters, $\omega$, to tune the model for an individual.

We note that there are two common approaches to training. One approach is to establish well-tuned learning rates via manual tuning or direct optimization~\cite{Hsieh2018OptimizingTL}. This approach ensures that the training of the neural network has both minimal convergence time and steady state error. A second option is to learn $\theta$ and $\omega$ by iteratively training one and then the other~ \cite{DBLP:journals/corr/LiH16e}. We utilize the former approach (i.e., concurrent training) due to empirical evidence in its favor.

\section{Bayesian Counterfactual Reasoning}
Counterfactual reasoning can be used to increase the power of machine learning algorithms\cite{foerster2018counterfactual}. Through pairwise comparisons between the actions taken and the set of actions not taken, a ranking formulation can be developed and used to predict which action the expert would ultimately take at each moment in time. \citet{Gombolay:2016a} presented evidence that learning a pairwise preference model by comparing pairs of actions can outperform a multi-class classification model. This paper is the first to our knowledge that has applied counterfactual reasoning in BNNs.

Our pairwise approach is similar to that of \citet{Gombolay:2016a} with the important distinction that we include Bayesian reasoning. For each observation, we receive a set of features for each action, $x_a^t$, as well as a set of state features, $\bar{x}^t$. For an observation in which action $a \in A$ was selected, we construct training example-label pairs $\tuple{x_{a,a'}^t,y_{a,a'}^t}$ according to Equations \ref{eq:pw1}-\ref{eq:pw4}. 
\begin{align}
x_{a,a'}^t &= [\omega,\bar{x}^t,x_a^t - x_{a'}^t] \label{eq:pw1} \\ 
y_{a,a'}^t &= 1 \label{eq:pw2} \\
x_{a',a}^t &= [\omega,\bar{x}^t,x_{a'}^t - x_a^t] \label{eq:pw3} \\ 
y_{a',a}^t &= 0 \label{eq:pw4}
\end{align} 
We generate these examples for all $a'$ in $A\backslash a$ and for all time observation time steps. After training a classifier, $f$, on this data set, we can predict the action selected by the user as shown in Equation \ref{eq:pairwise-for-scheduling_output}.
\begin{equation}
\hat{a}=\argmax_{a \in A} \sum_{a' \subset A} f (a, a',\omega)
\label{eq:pairwise-for-scheduling_output}
\end{equation}

\subsection{Bayesian Action Embeddings for Counterfactual Reasoning}
While the aforementioned approach to counterfactual reasoning is novel in application to BNNs for LfD, we develop an additional contribution demonstrating the capability of BNNs. One of the challenges with the approach in Equations \ref{eq:pw1}-\ref{eq:pw4} is that one must have access to action-specific features. In the context of our synthetic scheduling environment, the action features are readily available in the form of information about which agent is performing which task and the features of that task (e.g., it's location, deadline, duration, etc.). However, in our evaluation with real-world data from StarCraft II, such features are not directly provided.

In our evaluation of counterfactual reasoning in StarCraft II, we leverage BNNs to infer action embeddings by training a state transition model, $f$ to take as input a state, $s_t$ and an action embedding $\omega_a$, and predict the next state in a game's replay, $s_{t+1}$. By minimizing the R\'{e}nyi divergence (Equation ~\ref{eq:alpha} between $s_{t+1}$ and $f(s_t,\omega_a)$, for each action across all time steps and games, we can infer a player-invariant, action embedding, $\omega_a$ that can serve in lieu of $x_a^t$ in Equation \ref{eq:pw1} and \ref{eq:pw3}.

\subsection{Robust Bayesian Prediction}
One of the challenges with BNN-based prediction is that the model can provide highly-erroneous predictions at the beginning of a testing episode, when the Bayesian embedding has not yet converged from feedback. In the short-term, a baseline neural network can outperform a BNN. Yet, in the long-term, as we will see in our results section, a BNN will ultimately overtake the performance of baseline methods as the embedding converges and the BNN hones in on the desired, customized prediction model. To achieve the best of both worlds, we develop a hybrid methodology we denote $\textit{Hybrid-Bayes Shift}$.

We provide a pseudo-code description in Algorithm \ref{alg:switch}. We experimented with a number of criteria (e.g., determining whether the BNN method has a higher prediction accuracy over the previous $k$ time steps relative to the baseline method) but found that the best performance is achieved when the switch, between making predictions using the baseline method and the BNN, occurs after the embedding has converged, as described by Line 7 of Algorithm \ref{alg:switch}.



\begin{algorithm}[tb]
   \caption{Hybrid-Bayes Shift}
   \label{alg:switch}
\begin{algorithmic}[1]
   \STATE {\bfseries Input:} $\pi_{\theta,\omega}^{BNN}$, $\pi_{\phi}^{NN}$
   \STATE{$t \leftarrow 0$}
   \STATE{$ \pi \leftarrow \pi_{\phi}^{NN}$}
   \WHILE{episode not finished}
   \STATE $\hat{y}_t \leftarrow \pi(x_t)$
\STATE $\omega^{(t+1)} \leftarrow $ Apply Equation \ref{eq:BP_omega}
   \IF  {$\|\omega^{(t-1)}-\omega^{(t)}\| < \epsilon$}
   \STATE  $\pi \leftarrow \pi^{BNN}_{\theta,\omega}$
   \ENDIF
   \STATE $t \leftarrow t + 1$
   \ENDWHILE
\end{algorithmic}
\end{algorithm}

\section{Data Sets}
\label{datasets}
Two different data sets were used to validate our methodology, a synthetic scheduling environment and a real-world dataset of gameplay from StarCraft II.

\subsection{Scheduling Environment}
The first environment we use to explore BNNs for LfD is a synthetic environment that we can control, probe, and intuit to empirically validate the efficacy of our proposed method. The environment we develop is similar to one by \cite{Gombolay:2016a} developed for selecting scheduling actions (i.e., which agent to assign to which task) for a jobshop scheduling problem. The prior version of this benchmark employed one of three heuristic policies for selecting actions based upon the characteristics of the scheduling instance. These characteristics were made available to the observing LfD algorithm as to make the policy inference problem fully observable.

We adapt this environment by no longer providing our LfD algorithms the features necessary to tease out which heuristic would be applied for a given problem instance. Instead, the behavior appears heterogeneous and without clear rationale. For an LfD algorithm to properly handle this obfuscation, the algorithm must infer which policy is being employed based upon feedback from the actions being selected in real-time. We note that the dataset is not exclusively heterogeneous -- each policy adheres to basic scheduling constraints (e.g., no two agents can occupy the same physical location, etc.). As such, we expect even algorithms that assume homogeneity may have some success in cases in which only one or a few actions satisfy all scheduling constraints.

\subsection{StarCraft II}
A real-world data set with gameplay from StarCraft II is used to evaluate the utility of a personalized Bayesian embedding generated from a BNN. This data was provided alongside the StarCraft II API PySC2 \cite{vinyals2017starcraft}. This dataset contains a large number of 1-vs.-1 replays that affords access to game state-action information at every frame, information regarding the outcome of the game, and the ranking of the players. Each game can be thought as a demonstration trajectory by a human-human dyad.

The state of the game at any time step within the trajectory consists of several images pertaining to where units and buildings are located alongside information about visibility regions, and vectorized state information pertaining to the amount of resources, buildings, and units are in the game. The action taken in every frame can be one of hundreds, and thus as a simplification, we have produced forty actions that are representative of all. 


\section{Results and Discussion}
\label{results}
We assess the utility of inferring personalized Bayesian embeddings for LfD using the synthetic scheduling environment and StarCraft II replays described in Section \ref{datasets}.

\subsection{Synthetic Scheduling Environment}
We consider scheduling demonstrations in which a mock expert chose one of three heuristic policies to assign twenty tasks to two agents and sequence those tasks through time.
Several LfD architectures were compared to test their ability on a heterogeneous demonstration set and to further display the utility of applying a BNN to LfD. We compared a: 
\begin{itemize}
  \item neural network \`{a} la \citet{DBLP:conf/icml/SammutHKM92})
  \item set of k neural networks, each trained upon data clustered through k-means  \`{a} la \citet{Nikolaidis:2015:EML:2696454.2696455}
  \item a neural network where the data was first segmented using a Gaussian Mixture Model (GMM), and then input into a network augmented with its probability of coming from each segment
  \item Bayesian Neural Network
  \item Long-Short-Term-Memory (LSTM) network that can serve to learn temporal elements of the data
  \item Bayesian LSTM (B-LSTM), which is an LSTM with a Bayesian embedding in the input layer
\end{itemize}
Sample models of the networks and details regarding constraint satisfaction and data generation can be found on \url{https://github.com/ghost12331/HLfD}. Each of the methods described above were trained given 3, 9, 15, 150, and 1500 schedules of training data. 


\subsubsection{Standard Approach}
\begin{figure}[ht]
\begin{center}
\centerline{\includegraphics[width = \columnwidth]{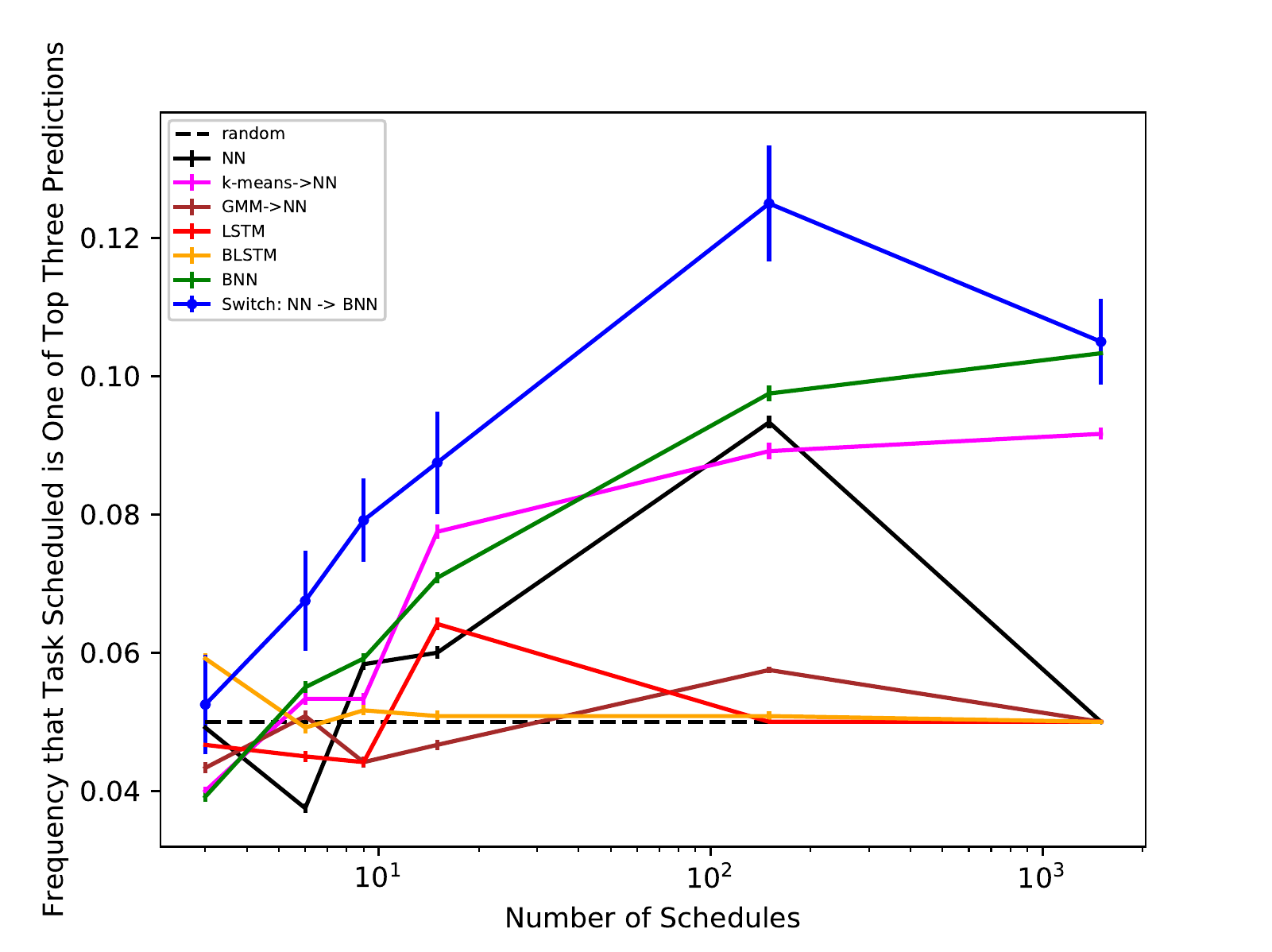}}
\caption{This figure depicts the prediction accuracy for the scheduling environment training to select the correct action.}
\label{fig:1_switch_results_naive_sched}
\end{center}
\end{figure}

The performance of each network architecture is shown in Figure \ref{fig:1_switch_results_naive_sched}, where the network is trained to take features of the state of the scheduling environment, including features describing each and every scheduling action, and map it to the action the mock expert would have selected. This figure demonstrates that our \emph{Hybrid-Bayes Shift} method (blue line in Figure \ref{fig:1_switch_results_naive_sched}) is able to outperform all baselines when the amount of data available for training is greater than three instances by explicitly reasoning about embedding convergence. 



\subsubsection{Bayesian Counterfactual Reasoning}
Next, we consider Bayesian counterfactual reasoning to increase the amount of information apparent in a given set, provide additional structure to the learning problem, and infer an embedding to represent the heterogeneity apparent in a given individual.

A slight difference between the procedure of using Bayesian counterfactual reasoning in the synthetic scheduling environment and StarCraft II is how the action specific features are obtained. In the scheduling environment, these features could be obtained directly from the data; however, in StarCraft, they could not.
Our approach to learn these embeddings is to utilize an action-BNN, where the latent embedding is a vector of action specific features. We learn the model $\pi_\psi : S_t \times \Omega_A \to S_{t+1}$, where $S_t$ is representative of the game state at time $t$, $S_{t+1}$ is the game state at time $t+1$, and $\Omega_A$ refers to a set of action-specific embeddings. After this model is learned, this set can be extracted and pairwise comparisons can be conducted between the set of actions taken, and not taken, as in Equations \ref{eq:pw1} - \ref{eq:pw4} with the exception that multiple actions may be taken.

\begin{figure}[ht]
	\begin{center}
		\centerline{\includegraphics[width = \columnwidth]{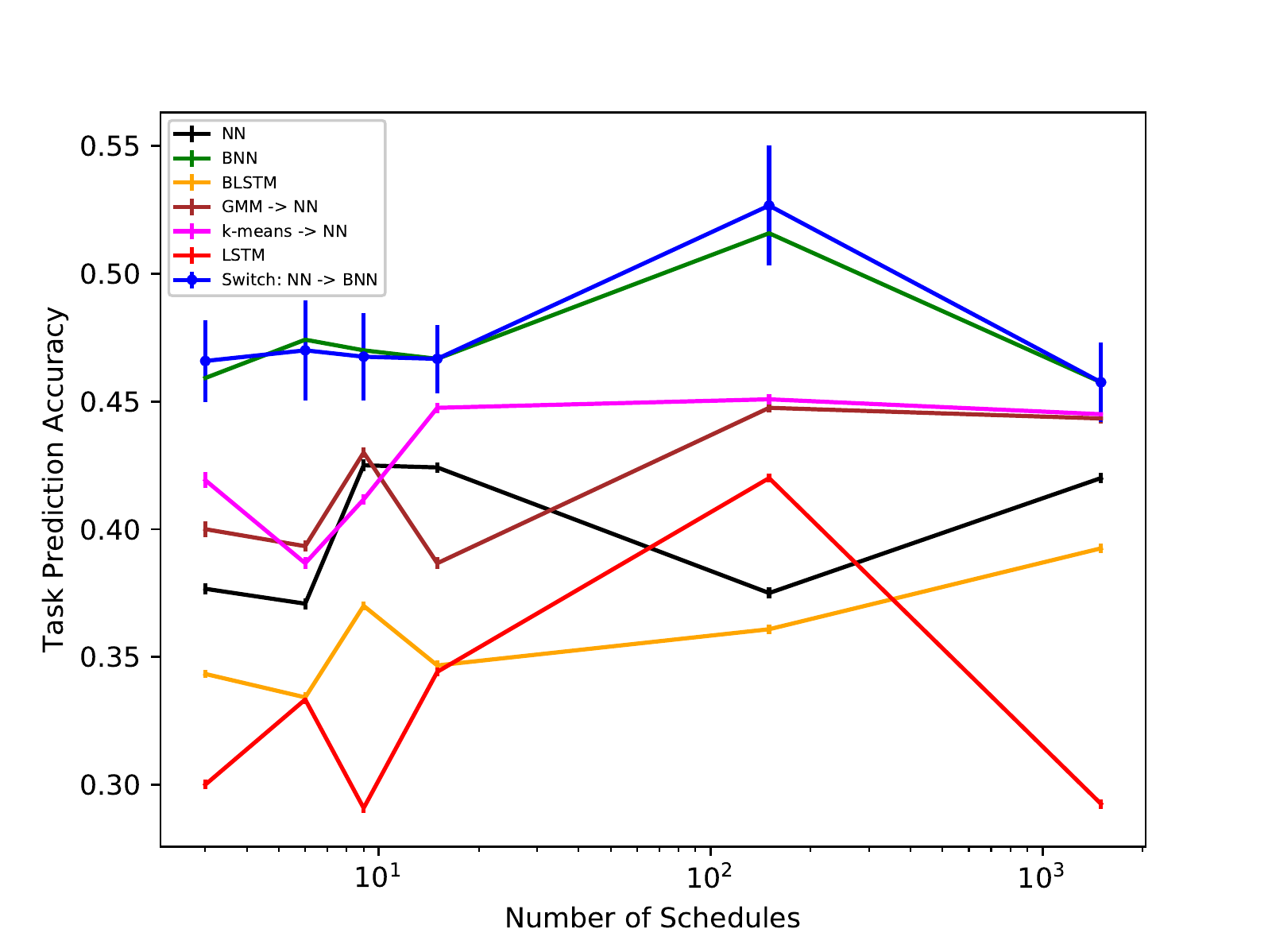}}
		\caption{This figure depicts the top-1 prediction accuracy with counterfactual reasoning for the scheduling environment }
		\label{fig:top1_switch_results_pairwise_sched}
	\end{center}
\end{figure}

Figure \ref{fig:top1_switch_results_pairwise_sched} provides evidence that  counterfactual reasoning greatly improves the power of our LfD models relative to the results from Figure \ref{fig:1_switch_results_naive_sched}. Utilizing counterfactual reasoning with Bayesian inferences allows for an approximately a 4-fold increase in the capability of the BNN to predict the task to schedule. As the BNN is again able to outperform conventional approaches in LfD, this provides further verification that inferring personalized Bayesian embeddings can aid in anticipatory models. With counterfactual reasoning, we can see that the BNN and the BNN-based \emph{Hybrid-Bayes Shift} method achieve approximately the same performance and outperform the method of \cite{Nikolaidis:2015:EML:2696454.2696455}. We can also see that the top-3 prediction accuracy for our method approaches $80\%$ and is strikingly robust to the amount of training data available, only varying from $75\%$ for three scheduling instances to a peak of $80\%$ for $150$ available demonstrations.


\begin{figure}[ht]
\begin{center}
\centerline{\includegraphics[width=\columnwidth]{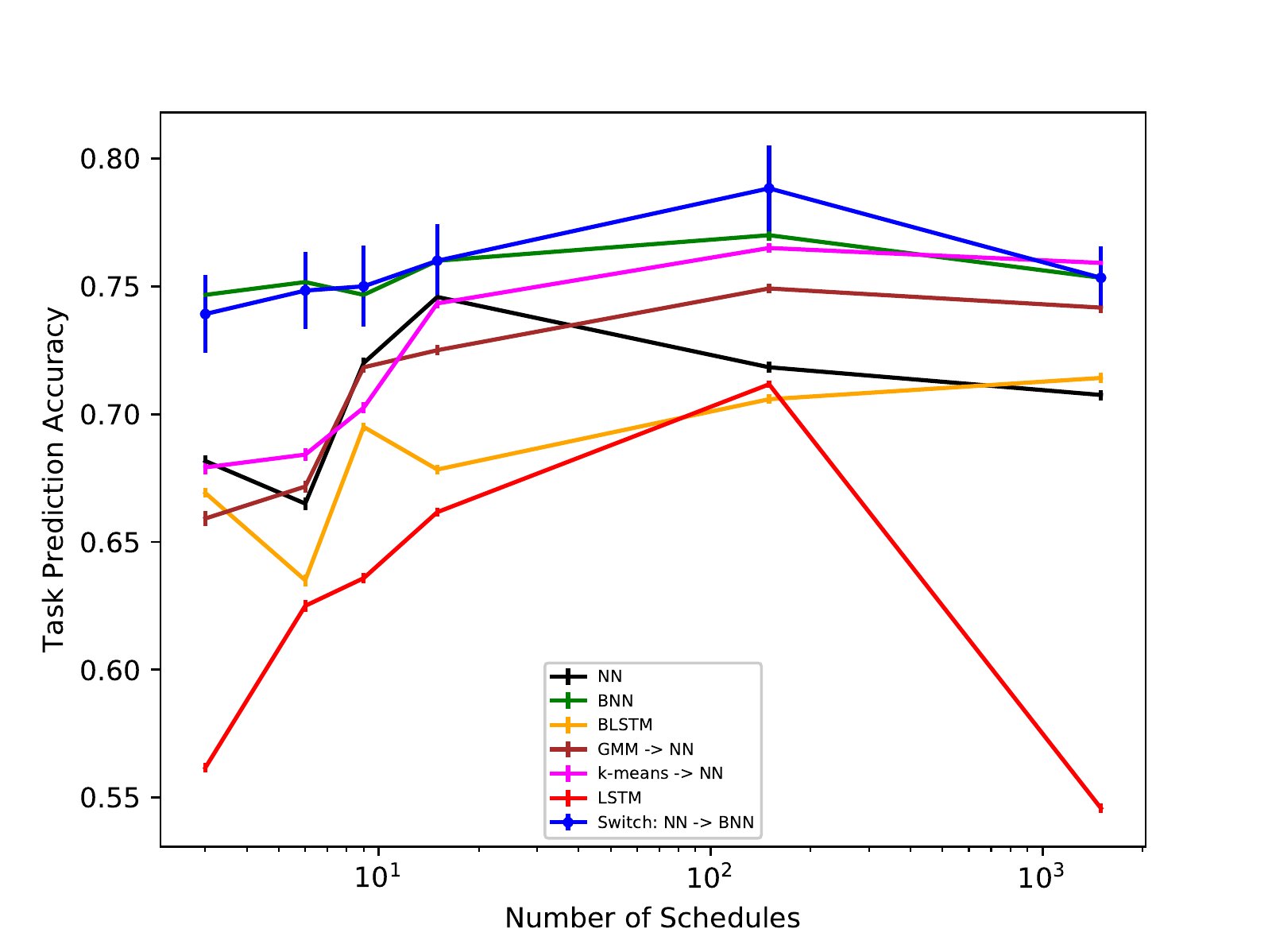}}
\caption{This figure depicts the top-3 prediction accuracy with counterfactual reasoning for the scheduling environment.}
\label{fig:top3_switch_results_pairwise_sched}
\end{center}
\end{figure}

%
%

\subsection{Real World Gameplay Data from StarCraft II}
Having shown promising results for the use of Bayesian-based neural networks, the \emph{Hybrid-Bayes Shift} method, and counterfactual reasoning, we seek to replicate our findings in a more complex domain with real-world data: StarCraft II. Specifically, we learn the gameplay policy of a given player playing StarCraft II. We again utilize the standard and counterfactual approach, and note the differences between the two methods. 
As our metric, we compare the performance of the ability to model a player policy by looking at the $\%$ difference in loss in comparison with a baseline neural network. Further information about the architecture or hyperparameters used can be found on \url{https://github.com/ghost12331/HLfD}.

\subsubsection{Standard Approach}
The performance of each method, BNN, B-LSTM, LSTM, and a baseline NN is shown in figure \label{Naive-Action}. We do not include the $GMM$ and $k-means$-based methods of \cite{Nikolaidis:2015:EML:2696454.2696455} as these were shown to under perform relative to our Bayesian approaches to LfD. For this policy learning problem, each model is given the current game-state and attempts to predict the action that will be taken at that timestep. 

\begin{figure}[ht]
\begin{center}
\centerline{\includegraphics[width = \columnwidth]{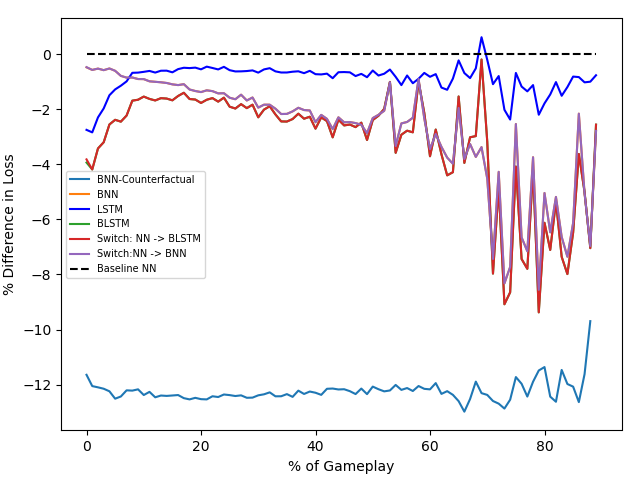}}
\caption{This figure depicts the loss for the StarCraft II environment with both counterfactual and non-counterfactual reasoning.}
\label{fig:Naive-Action}
\end{center}
\end{figure}

In Figure \ref{fig:Naive-Action}, we report the results of a BNN, LSTM, and B-LSTM trained using the standard, multi-class classification (i.e., not counterfactual) approach. We can see that the LSTM outperformed the baseline, NN, model by approximately $1\%$, which is to be expected given an LSTM's ability to learn a dynamic embedding passed from state to state. Interestingly, we found that a BNN -- without any recurrent structure except for its backpropagated embedding -- outperformed an LSTM model achieving an almost $10\%$ advantage over the baseline. We also found evidence that the B-LSTM performed as well as the BNN but with additional gains in early gameplay (i.e., the first $20\%$ of the game). These results provide strong evidence that the embedding learned for an individual player captures impactful information not easily recovered with a mere application of an LSTM.

\subsubsection{Bayesian Counterfactual Reasoning}
We report the performance of our action-embedding-learning method for counterfactual reasoning in BNNs in Figure \ref{fig:Naive-Action}. We can see that the BNN model trained via counterfactual reasoning (green curve in Figure \ref{fig:Naive-Action}) using Bayesian action embeddings achieves a factor of 3 reduction in the loss relative to a neural network baseline for the majority of the games' durations. These promising results show the potential of BNNs for adaptive learning from heterogeneous demonstration, BNNs for learning action embeddings in such domains, and BNNs for counterfactual reasoning.

\subsection{Discussion}
In our empirical evaluation, we provide evidence of the utility of learning personalized Bayesian embeddings for LfD tasks in two domains: one synthetic environment for jobshop scheduling and a virtual domain with real human gameplay data. Further, we show that counterfactual reasoning can provide a multi-fold improvement in performance. Finally, we demonstrate a simple, but powerful algorithm (Algorithm \ref{alg:switch}) can can compensate for the inaccuracy of a BNN's prediction before the embedding has converged. In the field of LfD, our approach achieves superior performance relative to state-of-the-art techniques~\cite{Nikolaidis:2015:EML:2696454.2696455}. 

\section{Limitations \& Future Work}
We also wanted to determine whether BNNs emergently produced embeddings that were person-specific but network-invariant. In other words, we ask whether a Bayesian embedding learned from one network can be fed into a different network -- perhaps one that was not able to leverage feedback to infer its own embedding online. Such a situation might arise from attempting to infer a player-specific value function over the course of a single game without knowing that game's outcome. One could train two networks, an action-prediction network and a value-prediction network. The action-prediction network could benefit from dense feedback (i.e., observing which action a player took at each time step) to infer an accurate embedding. The value-prediction network could likewise be trained on a point estimate of the expected reward, $V^{\pi_{\theta,\omega}}(s)\mathbb{E}[\sum_{t=0}^{\infty} \gamma^t r_t | s_o = s]$, when following a player-specific policy $\pi_{\theta,\omega}$ parameterized by network parameters $\theta$ and personalized embedding $\omega$. However, when attempting to predict the value of a new player in state $s$ during a real game (i.e., a ``cost-to-go'' given a spare reward structure assessed at the end of the game), one would not have the benefit of such feedback to be able to infer a Bayesian embedding. As such, one might seek to determine whether an embedding could be inferred via the action-network and passed to the value network.

\begin{figure}[ht]
\begin{center}
\centerline{\includegraphics[width = \columnwidth]{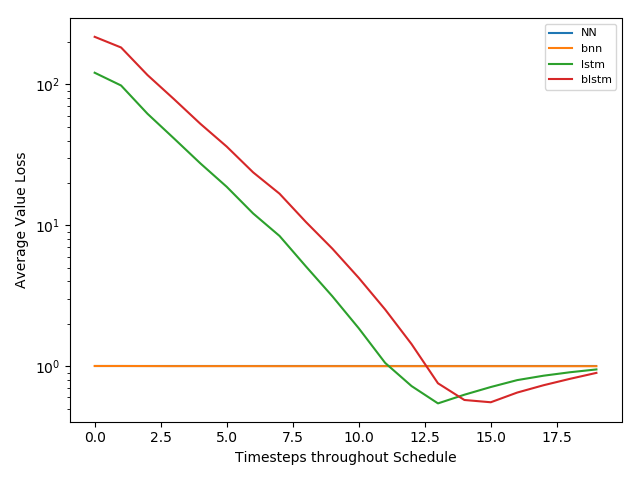}}
\caption{This figure depicts the results of a preliminary investigation into the emergence (or lack thereof) of network-invariant, personalized embeddings. The NN and BNN curves are overlapping.}
\label{fig:A2V}
\end{center}
\end{figure}

We conducted an experiment to investigate this possibility by employing a dense-feedback network to infer an embedding for a sparse-feedback network in the domain of StarCraft II. As shown in Figure \ref{fig:A2V}, our initial investigation did not yield positive results. This figure provides evidence that the B-LSTM value network actually performed worse than a non-Bayesian LSTM. We hypothesize that the LSTM is having to adapt its recurrent embedding to override the changing Bayesian input provided from the action network, which was a BNN.
These results are disappointing, suggesting that such an emergence of network-invariant, personalized embeddings is not guaranteed to happen. This result is, perhaps, unsurprising as the embedding is not being tailored to directly reduce the loss of the value network. In future work, we will explore the theoretical and algorithmic conditions for which such an emergence could happen. For example, one approach may be to adopt a meta-learning approach in which one seeks to repeatedly learn a Bayesian embedding for a person such that the distance between the converged embeddings, and across networks, at the end of an episode is minimized and the distance between the embeddings at the beginning of the episode is maximized. Such an approach could compel the networks to learn an intrinsic mechanism to relate persons in the way that generative adversarial networks provide evidence of embeddings amenable to vector addition and subtraction~\cite{radford2015unsupervised}.

\section{Conclusion}
In this paper, we investigated a novel formalism for learning from heterogeneous demonstration via BNNs. We showed that BNNs can be leveraged to infer personalized, latent embeddings to enable the network to better adapt to the unique characteristics of an individual's decision-making behavior. We also developed a Bayesian counterfactual approach that greatly increases the power of these networks for small data set with well-defined action descriptors. Finally, we proposed future work in learning personalized, network-independent embeddings to facilitate cross-network embedding sharing. Our results on synthetic and real-world data sets show a strong improvement over state-of-the-art baselines for learning from heterogeneous demonstration.

\balance

\bibliographystyle{named}
\bibliography{ijcai19}

\begin{thebibliography}{}

\bibitem[\protect\citeauthoryear{Abbeel and
  Ng}{2004}]{Abbeel:2004:ALV:1015330.1015430}
Pieter Abbeel and Andrew~Y. Ng.
\newblock Apprenticeship learning via inverse reinforcement learning.
\newblock In {\em Proceedings of the Twenty-first International Conference on
  Machine Learning}, ICML '04, pages 1--, New York, NY, USA, 2004. ACM.

\bibitem[\protect\citeauthoryear{Amershi \bgroup \em et al.\egroup
  }{2014}]{amershi2014power}
Saleema Amershi, Maya Cakmak, William~Bradley Knox, and Todd Kulesza.
\newblock Power to the people: The role of humans in interactive machine
  learning.
\newblock {\em AI Magazine}, 35(4):105--120, 2014.

\bibitem[\protect\citeauthoryear{Chernova and
  Veloso}{2007}]{chernova2007confidence}
Sonia Chernova and Manuela Veloso.
\newblock Confidence-based policy learning from demonstration using gaussian
  mixture models.
\newblock In {\em Proceedings of the 6th international joint conference on
  Autonomous agents and multiagent systems}, page 233. ACM, 2007.

\bibitem[\protect\citeauthoryear{Coates \bgroup \em et al.\egroup
  }{2009}]{coates2009apprenticeship}
Adam Coates, Pieter Abbeel, and Andrew~Y Ng.
\newblock Apprenticeship learning for helicopter control.
\newblock {\em Communications of the ACM}, 52(7):97--105, 2009.

\bibitem[\protect\citeauthoryear{Foerster \bgroup \em et al.\egroup
  }{2018}]{foerster2018counterfactual}
Jakob~N Foerster, Gregory Farquhar, Triantafyllos Afouras, Nantas Nardelli, and
  Shimon Whiteson.
\newblock Counterfactual multi-agent policy gradients.
\newblock In {\em Thirty-Second AAAI Conference on Artificial Intelligence},
  2018.

\bibitem[\protect\citeauthoryear{Gombolay \bgroup \em et al.\egroup
  }{2016}]{Gombolay:2016a}
Matthew Gombolay, Reed Jensen, Jessica Stigile, Sung-Hyun Son, and Julie Shah.
\newblock Decision-making authority, team efficiency and human worker
  satisfaction in mixed human-robot teams.
\newblock In {\em Proceedings of the International Joint Conference on
  Artificial Intelligence ({IJCAI})}, New York City, NY, U.S.A., July 9-15
  2016.

\bibitem[\protect\citeauthoryear{Hsieh and
  Shanechi}{2018}]{Hsieh2018OptimizingTL}
Han-Lin Hsieh and Maryam~M. Shanechi.
\newblock Optimizing the learning rate for adaptive estimation of neural
  encoding models.
\newblock In {\em PLoS Computational Biology}, 2018.

\bibitem[\protect\citeauthoryear{Killian \bgroup \em et al.\egroup
  }{2017}]{DBLP:conf/aaai/KillianKD17}
Taylor~W Killian, Samuel Daulton, George Konidaris, and Finale Doshi-Velez.
\newblock Robust and efficient transfer learning with hidden parameter markov
  decision processes.
\newblock In I.~Guyon, U.~V. Luxburg, S.~Bengio, H.~Wallach, R.~Fergus,
  S.~Vishwanathan, and R.~Garnett, editors, {\em Advances in Neural Information
  Processing Systems 30}, pages 6250--6261. Curran Associates, Inc., 2017.

\bibitem[\protect\citeauthoryear{Konidaris \bgroup \em et al.\egroup
  }{2011}]{Konidaris11a}
G.D. Konidaris, S.~Osentoski, and P.S. Thomas.
\newblock Value function approximation in reinforcement learning using the
  {F}ourier basis.
\newblock In {\em Proceedings of the Twenty-Fifth Conference on Artificial
  Intelligence}, pages 380--385, August 2011.

\bibitem[\protect\citeauthoryear{Konidaris \bgroup \em et al.\egroup
  }{2012}]{Konidaris12a}
G.D. Konidaris, S.R. Kuindersma, R.A. Grupen, and A.G. Barto.
\newblock Robot learning from demonstration by constructing skill trees.
\newblock {\em International Journal of Robotics Research}, 31(3):360--375,
  March 2012.

\bibitem[\protect\citeauthoryear{Lee \bgroup \em et al.\egroup
  }{2015}]{lee2015learning}
Alex~X Lee, Henry Lu, Abhishek Gupta, Sergey Levine, and Pieter Abbeel.
\newblock Learning force-based manipulation of deformable objects from multiple
  demonstrations.
\newblock In {\em 2015 IEEE International Conference on Robotics and Automation
  (ICRA)}, pages 177--184. IEEE, 2015.

\bibitem[\protect\citeauthoryear{Li and
  Hoiem}{2016}]{DBLP:journals/corr/LiH16e}
Zhizhong Li and Derek Hoiem.
\newblock Learning without forgetting.
\newblock {\em CoRR}, abs/1606.09282, 2016.

\bibitem[\protect\citeauthoryear{Mohammed \bgroup \em et al.\egroup
  }{2010}]{doi:10.1177/0149206309356804}
Susan Mohammed, Lori Ferzandi, and Katherine Hamilton.
\newblock Metaphor no more: A 15-year review of the team mental model
  construct.
\newblock {\em Journal of Management}, 36(4):876--910, 2010.

\bibitem[\protect\citeauthoryear{Nikolaidis and
  Shah}{2012}]{nikolaidis2012human}
Stefanos Nikolaidis and Julie Shah.
\newblock Human-robot teaming using shared mental models.
\newblock {\em ACM/IEEE HRI}, 2012.

\bibitem[\protect\citeauthoryear{Nikolaidis \bgroup \em et al.\egroup
  }{2015}]{Nikolaidis:2015:EML:2696454.2696455}
Stefanos Nikolaidis, Ramya Ramakrishnan, Keren Gu, and Julie Shah.
\newblock Efficient model learning from joint-action demonstrations for
  human-robot collaborative tasks.
\newblock In {\em Proceedings of the Tenth Annual ACM/IEEE International
  Conference on Human-Robot Interaction}, HRI '15, pages 189--196, New York,
  NY, USA, 2015. ACM.

\bibitem[\protect\citeauthoryear{Odom \bgroup \em et al.\egroup
  }{2015}]{10.1007/978-3-319-19551-3_26}
Phillip Odom, Vishal Bangera, Tushar Khot, David Page, and Sriraam Natarajan.
\newblock Extracting adverse drug events from text using human advice.
\newblock In John~H. Holmes, Riccardo Bellazzi, Lucia Sacchi, and Niels Peek,
  editors, {\em Artificial Intelligence in Medicine}, pages 195--204, Cham,
  2015. Springer International Publishing.

\bibitem[\protect\citeauthoryear{Radford \bgroup \em et al.\egroup
  }{2015}]{radford2015unsupervised}
Alec Radford, Luke Metz, and Soumith Chintala.
\newblock Unsupervised representation learning with deep convolutional
  generative adversarial networks.
\newblock {\em arXiv preprint arXiv:1511.06434}, 2015.

\bibitem[\protect\citeauthoryear{Sammut \bgroup \em et al.\egroup
  }{2002}]{DBLP:conf/icml/SammutHKM92}
Claude Sammut, Scott Hurst, Dana Kedzier, and Donald Michie.
\newblock Imitation in animals and artifacts.
\newblock pages 171--189, 2002.

\bibitem[\protect\citeauthoryear{Terrell and
  Mutlu}{2012}]{Terrell:2012:RAM:2392800.2392849}
Allison Terrell and Bilge Mutlu.
\newblock A regression-based approach to modeling addressee backchannels.
\newblock In {\em Proceedings of the 13th Annual Meeting of the Special
  Interest Group on Discourse and Dialogue}, SIGDIAL '12, pages 280--289,
  Stroudsburg, PA, USA, 2012. Association for Computational Linguistics.

\bibitem[\protect\citeauthoryear{Vinyals \bgroup \em et al.\egroup
  }{2017}]{vinyals2017starcraft}
Oriol Vinyals, Timo Ewalds, Sergey Bartunov, Petko Georgiev, Alexander~Sasha
  Vezhnevets, Michelle Yeo, Alireza Makhzani, Heinrich Küttler, John Agapiou,
  Julian Schrittwieser, John Quan, Stephen Gaffney, Stig Petersen, Karen
  Simonyan, Tom Schaul, Hado van Hasselt, David Silver, Timothy Lillicrap,
  Kevin Calderone, Paul Keet, Anthony Brunasso, David Lawrence, Anders Ekermo,
  Jacob Repp, and Rodney Tsing.
\newblock Starcraft ii: A new challenge for reinforcement learning, 2017.

\bibitem[\protect\citeauthoryear{Ziebart \bgroup \em et al.\egroup
  }{2008a}]{Ziebart:2008:MEI:1620270.1620297}
Brian~D. Ziebart, Andrew Maas, J.~Andrew Bagnell, and Anind~K. Dey.
\newblock Maximum entropy inverse reinforcement learning.
\newblock In {\em Proceedings of the 23rd National Conference on Artificial
  Intelligence - Volume 3}, AAAI'08, pages 1433--1438. AAAI Press, 2008.

\bibitem[\protect\citeauthoryear{Ziebart \bgroup \em et al.\egroup
  }{2008b}]{ziebart2008maximum}
Brian~D Ziebart, Andrew~L Maas, J~Andrew Bagnell, and Anind~K Dey.
\newblock Maximum entropy inverse reinforcement learning.
\newblock In {\em Aaai}, volume~8, pages 1433--1438. Chicago, IL, USA, 2008.

\bibitem[\protect\citeauthoryear{{\.Z}yczkowski}{2003}]{zyczkowski2003renyi}
Karol {\.Z}yczkowski.
\newblock R{\'e}nyi extrapolation of shannon entropy.
\newblock {\em Open Systems \& Information Dynamics}, 10(03):297--310, 2003.

\end{thebibliography}

\end{document}